\documentclass[letterpaper, 10 pt, conference]{ieeeconf}  % Comment this line out if you need a4paper

\IEEEoverridecommandlockouts                              % This command is only needed if 
\usepackage{cite}
\usepackage{amsmath,amssymb,amsfonts}
\usepackage{algorithmic}
\usepackage{graphicx}
\usepackage{textcomp}
\usepackage{xcolor}
\def\BibTeX{{\rm B\kern-.05em{\sc i\kern-.025em b}\kern-.08em
    T\kern-.1667em\lower.7ex\hbox{E}\kern-.125emX}}
    
\title{\LARGE \bf
A Dataset and System for Real-Time Gun Detection in Surveillance Video Using Deep Learning}

\author{Delong Qi$^{1}$, Weijun Tan$^{1,2}$,    
    Zhifu Liu$^{1}$, Qi Yao$^{1}$, and Jingfeng Liu$^{1}$ % <-this% stops a space

\thanks{$^{1}$Delong Qi, Zhifu Liu, Qi Yao and Jingfeng Liu are with Shenzhen Deepcam Information Technologies, China
        {\tt\small \{delong.qi,zhifu.liu,qi.yao,jingfeng.liu\} @deepcam.com}}%
\thanks{$^{2}$Weijun Tan is with LinkSprite Technologies USA, and Shenzhen Deepcam Information Technologies, China
        {\tt\small weijun.tan@linksprite.com}}%
}

\begin{document}

\maketitle
\thispagestyle{empty}
\pagestyle{empty}

\begin{abstract}
 Gun violence is a severe problem in the world, particularly in the United States. Deep learning methods have been studied to detect guns in surveillance video cameras or smart IP cameras and to send a real-time alert to security personals. One problem for the development of gun detection algorithms is the lack of large public datasets. In this work, we first publish a dataset with 51K annotated gun images for gun detection and other 51K cropped gun chip images for gun classification we collect from a few different sources. To our knowledge, this is the largest dataset for the study of gun detection. This dataset can be downloaded at www.linksprite.com/gun-detection-datasets. We present a gun detection system using a smart IP camera as an embedded edge device, and a cloud server as a manager for device, data, alert, and to further reduce the false positive rate. We study to find solutions for gun detection in an embedded device, and for gun classification on the edge device and the cloud server. This edge/cloud framework makes the deployment of gun detection in the real world possible.  
 \end{abstract}

%\begin{IEEEkeywords}
%Gun detection, dataset, real-time, embedded device, classification, CNN   
%\end{IEEEkeywords}

%%%%%%%%% BODY TEXT
\section{Introduction}

Gun violence has been a big problem around the world, especially in the countries where it is legal to carry a gun, like the United States. Every year, many innocent people are killed or harmed by gun violence in public areas \cite{Gun-death-stats}.

There are video surveillance systems deployed in many public places, but they still require the supervision and intervention of humans. As a result, these system cannot detect gun-related crimes fast enough to prevent the crimes. In more recent years, automatic gun detection systems have been studied \cite{Grega,handgun-FasterRCNN,Olmos,Olmos2,Lim,Real-time-CCTV,CS231,posegun,e2e}. Most of them still use surveillance video cameras, and the video streams are sent to an edge device or server to decode the video, detect guns in the video frame, and send alert (email, text message, or phone call) to personnel on duty. Others use affordable smart IP camera as an edge device, do gun detection on the camera, send the cropped gun image to server, and activate the server to send alert. The former can reuse existing surveillance video cameras, but the video streaming uses a lot of WiFi or Ethernet bandwidth, and the video processing and gun detection use expensive edge devices or servers.  The latter needs to deploy new smart IP cameras, but the cameras accomplish all the video processing and gun detection, only very little data is sent to server for further processing. As a result, it is a more affordable solution than the former one.  

There are a few challenges in gun detection system. First of all, it has to achieve a very high true positive (correct detection) rate (TPR), and very low false positive (false alarm) rate (FPR). In our study, we find that one major problem is how to control the false alarms, i.e., the FPR. In this prospective, gun detection remains an open problem.   

The second challenge is there are no large public gun detection datasets.  Almost all all researchers collect their own dataset and evaluate the performance of their algorithms on these private dataset. To our knowledge, the amount of data in their dataset is typically a few thousands, e.g.,\cite{Olmos,Lim}, which is way too few.  As a result, there is no way researchers can compare the performance of their algorithms to others and it is impossible to tell which algorithm is better.  Our first contribution of this work is that we collect a dataset with 51K gun images. For the problem of false alarm we mentioned before, we also collect a non-gun image dataset, including images that are easily mis-detected as guns, for the study of gun/non-gun classification.  

Our second contribution is to present a system for real-time gun detection and reporting. We present the image processing procedures on the edge device and the cloud server.  We study to find solution for real-time gun detection on low-cost embedded edge device (e.g.,smart IP camera),  solution for light-weight gun classification on the edge device, and solution for more complex gun classification on the cloud server.  Our focus is not to invent new detection algorithm, but to find and optimize existing algorithms for use in a practical system. 

%------------------------------------------------------------------------

\section{Related Work}

Among the works on gun detection \cite{Grega,handgun-FasterRCNN,Olmos,Olmos2,Lim,Real-time-CCTV,CS231}, most of them focus on applying existing CNN networks on object detection of guns. They use very complex CNN networks without considering the applications in practical system. Even though some of them call their algorithms real-time \cite{Real-time-CCTV, CS231}, these algorithms use very expensive GPU machine or CPU machine, which make them not practical in large scale deployment. The work \cite{posegun} combines body pose with object detection to improve the detection accuracy. This additional pose detection makes it run even slower on given machine. 

Another problem is the lack of large public dataset for gun detection. Most of the works use proprietary dataset, which are not only small, but also not available to other researchers.  We summarize the CNN networks and datasets used in their work in Table \ref{T0}. Since our goal is to find a solution for embedded device, we add comments if these networks are applicable in practical systems.  From the summary, we observe that people really like Faster-RCNN \cite{FasterRCNN}, which is way more complex and slow for use in practical system.  The one exception is \cite{Grega}, which uses a transitional image processing and a 3-layer MLP. However, it uses a very small dataset of only 3559 frame images.  From this summary we see the need of a large dataset and a detection network for practical real-time applications.  

\begin{table}[ht]
	 \centering
	 \caption{CNN networks and datasets used in existing works}
	 \label{T0}
	 \begin{tabular}{cccc}
	    \hline
	    Ref & Method & dataset & comments\\
	    \hline
	    \cite{Grega} & MLP & 3559 & edge + 3-layer MLP \\
	    \cite{handgun-FasterRCNN} & fasterRCNN & - & Complex and slow\\
	    \cite{Olmos,Olmos2} & fasterRCNN & 8996 & Complex and slow\\
	    \cite{Lim} & M2Det & 5000 & Complex and slow\\
	    \cite{Real-time-CCTV} & FasterRCNN & - & Complex and slow\\
	    \cite{CS231} & VGG16 etc. & 3500 & Complex and slow\\
	    \cite{posegun} & YoloV3+Pose & 1745 & Complex \\
	    \hline
	    \textbf{Ours} & \textbf{Simplified CenterNet} & \textbf{51K} & \textbf{Real-time on camera} \\
	    \hline
	  \end{tabular}
\end{table}

\section{Gun Detection Dataset}
Before we present our gun dection system, we first present the dataset we prepare and make public for the community. 

\subsection{Image Sources}

The sources where we collect our gun image dataset include:

\begin{itemize}
  \item We first collect a lot of gun images from the IMFDB website \cite{IMFDB} - a movie internet firearms database. Then we use a CNN-based gun detector to roughly label the data. Finally we manually check and relabel the inaccurate labels. 
  \item We collect some images from publicly available websites of some of the papers \cite{Olmos,Lim,UCF}.  Note that many of these images overlap with the first source and we have to manually clean them. The images from the video footages in \cite{UCF} are used in \cite{Lim}. We do not use most of them because their resolution is very low and different from real cameras.  
  \item We deploy a few cameras in a private office and capture some gun detection data in a real scene. The data we collect this way is about a couple of thousands. 
  \item Many negative samples for gun detection and classification are collected from the failed cases in our deployment cameras.    
\end{itemize}

For the concern of FPR, we also collect a large non-gun image dataset for detection and classification. Since they are non-guns, so no annotation is needed. Non-gun images are rather abundant and easy to collect.  We collect 94K non-gun images from the popular ImageNet \cite{Imagenet} and COCO \cite{COCO} datasets. Furthermore, we collect objects falsely detected as guns in our system and put them in the dataset. These are hard cases and are very helpful to improve the FPR.    

\begin{figure*}[t]
    \centering
    \includegraphics[scale=0.62]{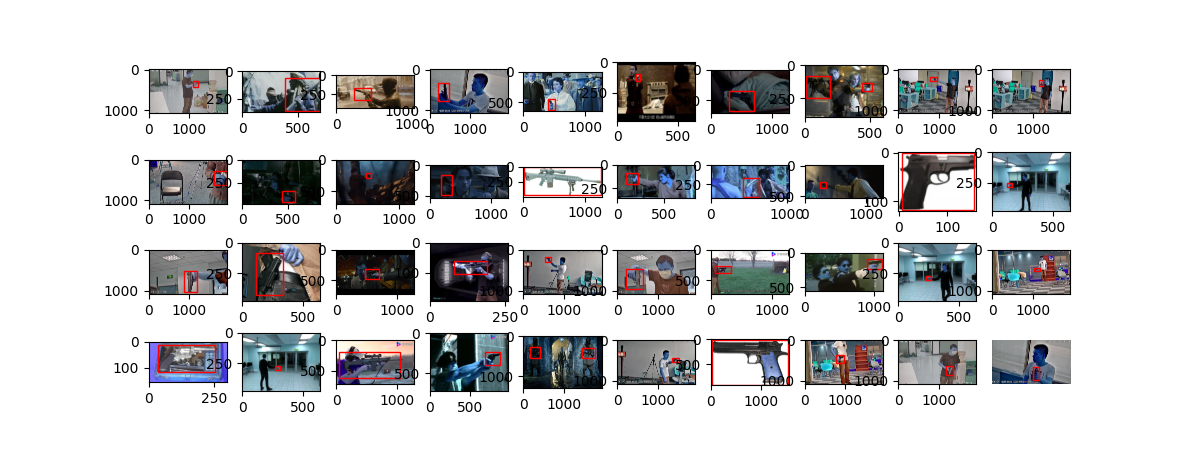}
    \caption{Sample gun detection images}
    \label{fig1}
\end{figure*}

\begin{figure*}[t]
    \centering
    \includegraphics[scale=0.76]{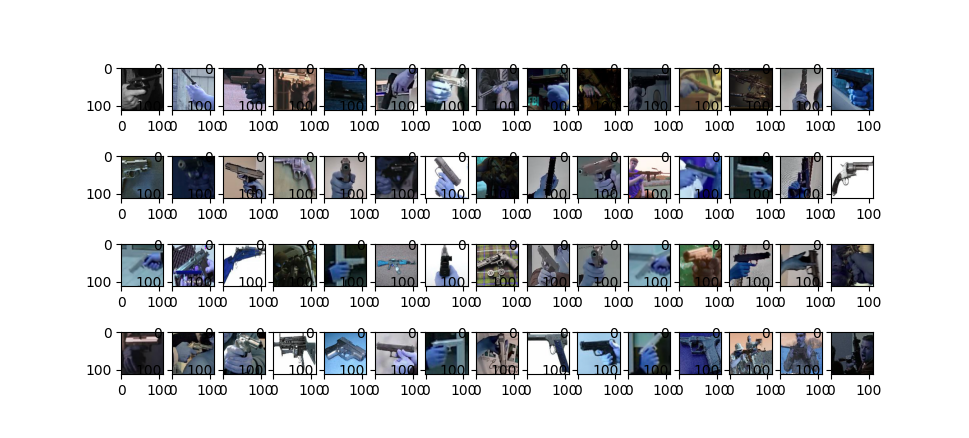}
    \caption{Sample gun classification images}
    \label{fig2}
\end{figure*}

In Fig.\ref{fig1}, we show 40 sample gun detection images where annotated gun bounding boxes are drew with red rectangles.  In Fig.\ref{fig2}, we show 40 sample gun classifier images, which are cropped from the ground truth locations of the some gun detection images.       

\subsection{Dataset Folder Structure}

\begin{figure}[t]
    \centering
    \includegraphics[scale=1.0]{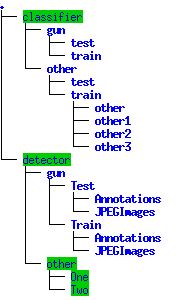}
    \caption{Dataset folder structure}
    \label{fig3}
\end{figure}

The dataset folder structure is shown in Figure~\ref{fig3}. In the top dataset folder, there are two subfolders - "detector" and "classifier", where all images for the gun detection and classification are saved. Any folder named "gun" is a folder where gun images are saved, and any folder named "other" is a folder where non-gun images are saved. In the "detector" folder, we split images to training set and test set. Users are free to re-split the training and test sets. Please note that this folder structure is the one we have when we decide to publish the dataset. The split of training and test dataset is not necessarily the one we use in our performance benchmark.     

The number of images are summarized in Table \ref{T2}. Please note that many of the negative samples of both detector and classification are from failed cases in our study. In the gun images, majority are handgun, while a good number of them are rifles, machine guns and other types.  

\begin{table}[ht]
	 \centering
	 \caption{Number of images}
	 \label{T2}
	 \begin{tabular}{ccc}
	    \hline
	    Model & Gun/Non-Gun(Other) & Number\\
	    \hline
	    Detector & Gun & 51,889 \\
	    Detector & Other  & 154798 \\
	    \hline
	    Classifier & Gun  &  51,398 \\
	    Classifier & Other & 371,078 \\
	    \hline
	    \end{tabular}
\end{table}

\subsection{Annotation File Format}

In the gun folders for detection, there are two sub-folders: JpegImages, and Annotations. All gun images in JPG format are saved in the JpegImages folder, and their corresponding annotation files are saved in the folder Annotations. The file names, one for JPG, and another for the annotation, are one to one mapped in the two folders. 

\begin{figure}[t]
    \centering
    \includegraphics[scale=0.75]{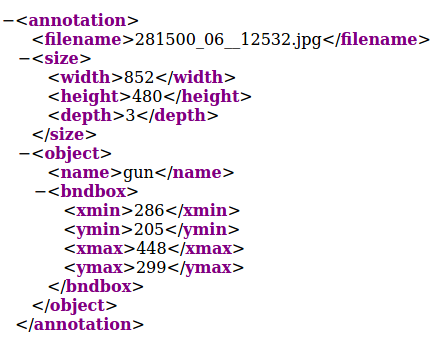}
    \caption{XML annotation format}
    \label{fig4}
\end{figure}

The annotation file is in XML format. In the annotation file, the filename, dimension of the whole image, and the upper-left and lower-right coordinates of guns are given.   An example of such a file is shown in Fig.\ref{fig4}. 

\begin{figure}[t]
    \centering
    \includegraphics[scale=0.6]{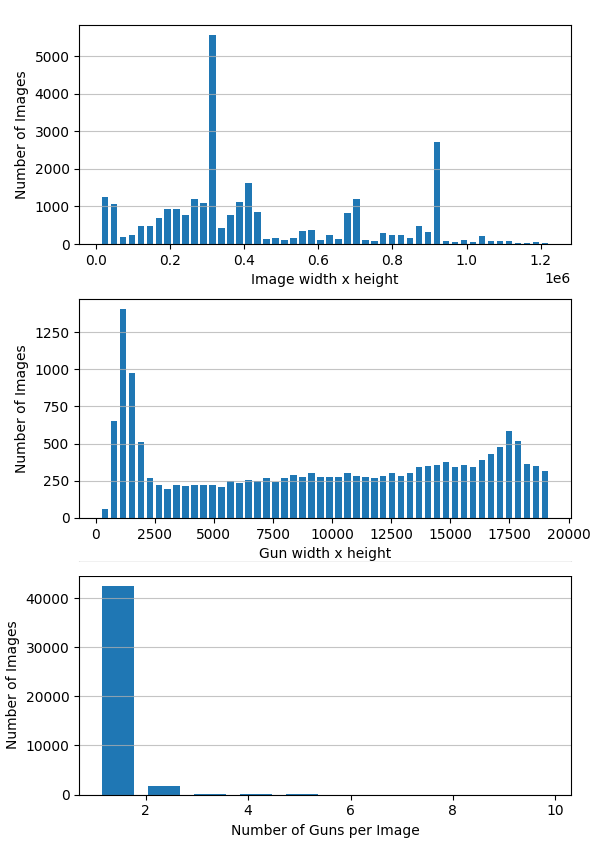}
    \caption{Gun detection image statistics}
    \label{fig5}
\end{figure}

\subsection{Statistics of Gun Images}

We check the statistics of the gun images in the detector training set. We look at the whole image size (width x height) and the gun bounding box size (width x height). These area indicates the resolution of the whole image and the gun object. We also look at the number of gun objects in every image.  Shown in Fig.\ref{fig5} are the statistics. 

From the distribution, we notice that 1) there are two peaks in the whole image size, which correspond to the widely used image sizes in surveillance video cameras and IP cameras. There are also a very few very large images that are not shown in the distribution; 2) The sizes of the gun objects are pretty much uniform distributed, except for the very small sizes. This imposes an challenge in gun detection because there exist a lot of very small gun objects in images; 3) There are mostly one gun object in every image. Some images have two gun objects, and very few images have up to 10 gun objects in an image.         

In the gun classification training dataset, there is always one gun object in every image. These images are all resized to 112x112.  

\section{Gun Detection System}

Our gun detection system consists of edge devices - which are mostly smart cameras and the cloud server. The edge device does most of the video processing, gun detection, first-level gun/non-gun classification, sending gun chip image or frame to server.  The server manages the device, data, alert.  In addition, it has second-level gun/non-gun classification, and extra level gun action recognition, which is still in progress.

\subsection{Gun Detection Process on Edge Device}

We use low-cost smart camera as edge device. A typical example is one uses the HiSilicon-Hi3516 chip set, or the Rockchip-RK3299 chip set. We do not consider traditional IP camera since it does not have the capability to run object detection using deep learning. Streaming the video of IP camera to an expensive edge device or server is not preferred.  

\begin{figure}[t]
    \centering
    \includegraphics[scale=0.4]{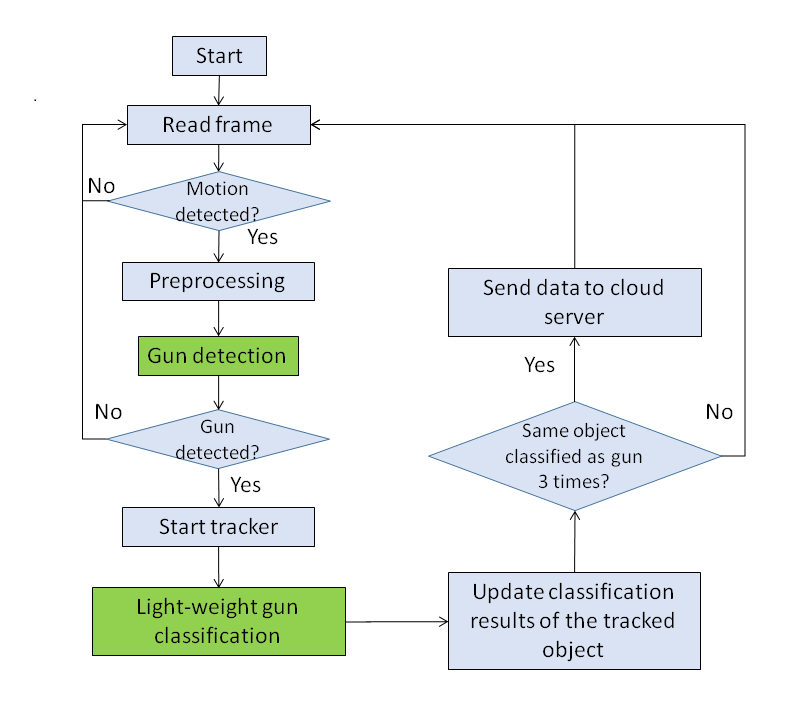}
    \caption{Gun detection process on edge device}
    \label{fig6}
\end{figure}

The gun detection process on edge device is similar to \cite{e2e}. The gun detector directly takes the original high resolution video stream as input. It first runs a motion detection to determine whether there are moving objects in the video to prevent the false detection in static scene. Once motion is detected in video, gun detection is conducted on a smaller resolution video. If a gun is detected, an object tracker is activated to track its movement. In the process of tracking, the light-weight classification network starts to run. Finally, if the tracked object is classified as a gun for more than three times, it is determined to be a gun. This tracker, classifier and some logical processing are added to minimize the false detection rate. After this, a cropped chip image of gun, a snapshot of the scene, and some Meta data such as device ID, bounding box of the gun, the timestamp, etc., are sent to the cloud, as shown in Fig.\ref{fig6}\cite{e2e}. 

\begin{figure}[t]
    \centering
    \includegraphics[scale=0.55]{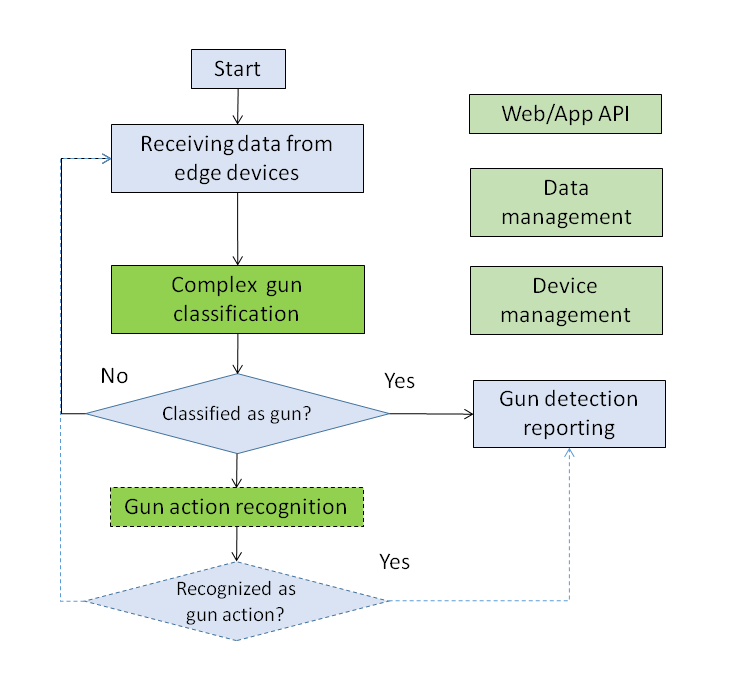}
    \caption{Gun classification process on cloud server.}
    \label{fig7}
\end{figure}

\subsection{Further Processing on the Cloud Server}

After receiving the data sent from the edge device, the server begins further processing the gun chip image and the snapshot scene image.  The function blocks of the server are shown in Fig. \ref{fig7}. 

First a second level of classification, which is a more complex model than the light-weight classifier on the edge device, is run. This classifier runs on the server, so it is not restricted to use expensive hardware resources like GPU.   

One extra level of function to control the FPR is to use gun action recognition. Since the server receives consecutive scene snapshots while the tracker is running, these snapshots carry temporal information of the motion of the gun and the person. These multiple scene snapshots can be used to do action recognition. Gun action is one type of abnormal action detection on videos \cite{UCF}. This work is still in progress, denoted by dashed lines in Fig. \ref{fig7}. For more details of the action recognition, the readers are referred to \cite{survey} and more in the literature.     

\begin{figure}[t]
    \centering
    \includegraphics[scale=0.6]{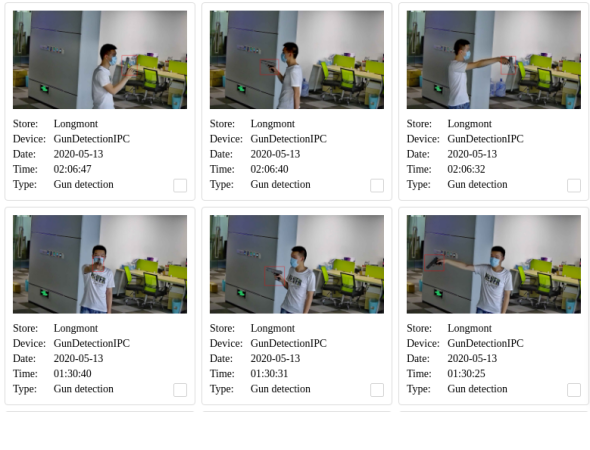}
    \caption{Demo of snapshot scene of detected gun.}
    \label{fig8}
\end{figure}

Other functions of the cloud server include management of device, data, reporting, and an API for web or mobile App. In the web server console, snapshot scene images of detected guns are displayed. A demonstration is shown in Fig. \ref{fig8}\cite{e2e}. 

\section{CNN Networks for Gun Detection and Classification}

In this section we present details of the CNN networks for gun detection and classification.  

\subsection{Gun Detection Network}

There are a lot of great progresses in object detection including the SSD \cite{SSD}, YOLO family \cite{YOLO3}, Faster-RCNN family \cite{FasterRCNN}, the feature pyramid network \cite{FPN}, MMDetection \cite{MMDetection}, EfficientDet \cite{EfficientDet}, transformer(DETR) \cite{DETR}, Centernet \cite{CenterNet,CenterNet2}, and so on.  Most of them have outstanding performance but run very slowly. For real-time object detection, the YOLO3, SSD, and Centernet are a few good choices. After extensive experiments, we find that the Centernet is a good choice offering a good trade-off between performance and speed. We do not present the results on other networks that we have studied.  

The original Centernet uses a few backbone networks, including the Hourglass-104, DLA-34, ResNet-101, and ResNet-18 \cite{CenterNet,CenterNet2}. We evaluate these choices on a HiSilicon-Hi3516 embedded platform, and find that their FPSs are all too slow for real time applications. Then we go further to test different light-weight backbone networks, including the MobileNet \cite{MobileNetV2} and VovNet \cite{Vovnet}. The results are shown in Table \ref{T3}. We see that these backbones can mostly meet the real-time or near real-time requirement.    

\begin{table}
	\begin{center}
		\caption{Different backbone network in the Centernet. The AP(0.5:0.05:0.95) performance is on the COCO-2017 val set}
		\label{T3}
		\begin{tabular}{ccc}
		    \hline
			Backbone & AP(\%) & FPS\\
			\hline
            MobileNetv2 & 32.2 & 12\\
            VoVNetv2-19-slim & 33.2 & 25\\
            VoVNetv2-19 & 35.1 & 10 \\
            VoVNetv2-39-slim & 36.1 & 20\\
            VoVNetv2-39 & 40.0 & 8\\
			\hline
		\end{tabular}
	\end{center}
\end{table}

Next we use the VoVNetv2-39 \cite{Vovnet} as example and study different data augmentation techniques, loss function, learning rate scheme etc. to improve its detection performance.  In these experiments, we use VOC2007+VOC2012 training sets as the training dataset, and use VOC2007 test set as the test dataset.  Listed in Table \ref{T4} are the experiment results. The best techniques are used in our gun detection performance evaluation.    

\begin{table}
	\begin{center}
		\caption{Techniques to improve the Centernet performance on VOC dataset. Training dataset = VOC2007+VOC2012 training sets, test set = VOC2007 test set.}
		\label{T4}
		\begin{tabular}{cc}
		    \hline
			Technique & mAP(\%) at IOU=0.5\\
			\hline
            basline & 74.1\\
            +multi-scale & 75.1\\
            +Giou loss \cite{giouloss} & 75.8 \\ 
            +Cosine LR & 76.1 \\ 
            +Mosaic \cite{yolo5} & 76.9 \\
            +FPN \cite{FPN},PAN \cite{PANet},RFB 
            \cite{RFB} & 78.1\\
			\hline
		\end{tabular}
	\end{center}
\end{table}

Finally we test the performance of the VoVNetv2-19-slim on our gun detection dataset. For comparison a couple of other networks are also evaluated.  The results are presented in Table \ref{T5}. The IOU threshold used is 0.3. We choose this threshold because we still have the gun/non-gun classifiers on the edge device and on the cloud server. Using a small IOU threshold will improve the recall rate of gun detection, while the classifier can filter out non-gun objects.     

\begin{table}
	\begin{center}
		\caption{The performance of the Centernet gun detection at IOU threshold = 0.3. Acc=accuracy, Rec=Recall, Pre=precision}
		\label{T5}
		\begin{tabular}{cccc}
		    \hline
			Backbone &  Acc \% & Rec \% & Pre \%\\
			\hline
			VoVNet-v2-19\_slim\_light  & 81.06 & 90.19 & 85.05\\
			VoVNet19\_slim & 84.28 & 90.33 & 88.60\\
			Resnet18 & 84.28 & 89.88 & 88.86\\
			\hline
		\end{tabular}
	\end{center}
\end{table}

\subsection{Gun Classification Network}

A light-weight gun/non-gun classifier runs on the edge device, and a more complex one runs on the cloud server. The one on the cloud sever is not restricted to use complex hardware like GPU. Therefore, we use a large classification network to ensure the best accuracy to filter out false positives detection.

\begin{figure}[t]
    \centering
    \includegraphics[scale=0.45]{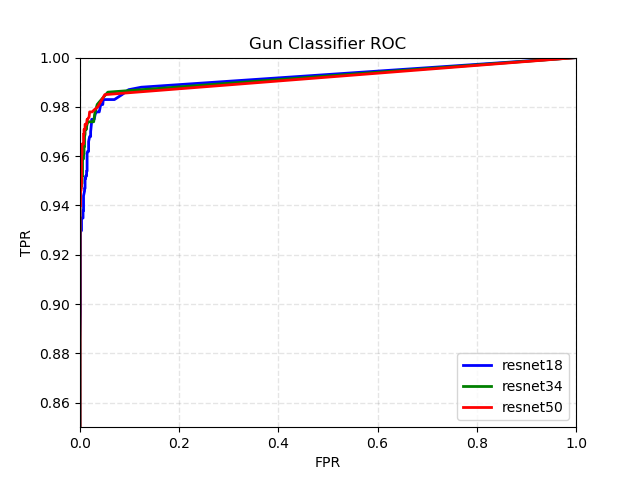}
    \caption{ROC of gun/non-gun classifiers}
    \label{roc}
\end{figure}

We choose to use Resnet \cite{Resnet}, with different depths and compression on number of channels. We train the classification network with 51K gun images, and 94K negative sample images. The test set includes 998 positive samples and 980 negative samples. The final test results are shown in Table \ref{T6}. We notice that even with Resnet50, the accuracy is only 97.83\%. This is because there are some very hard cases where non-gun objects are classified as guns. So this problem is still to be solved. Shown in Fig. \ref{roc} is the ROC curve of the classifier. Please note that, in practice we can use the detection and classification results in a few consecutive frames to improve the recall and precision. 

Based on these results, we choose to use the Resnet18 for the light-weight classifier on the edge device, and to use the Resnet50 for the complex classifier on the cloud server.  

\begin{table}[ht]
	 \centering
	 \caption{The performance of gun classifier at threshold = 0.5. Acc=accuracy, Rec=Recall, Pre=precision}
	 \label{T6}
	 \begin{tabular}{cccc}
	    \hline
	    Model & Acc(\%) & Rec(\%) & Pre(\%)\\
	    \hline
	    Resnet18 & 96.97 &	98.55 & 95.39\\
	    ResNet34 & 97.57 & 	99.27 & 95.89\\
	    Resnet50 &	97.83 &	99.69 & 95.99\\
	    \hline
	 \end{tabular}
\end{table}

\section{Conclusion}

We collect and publish a gun detection and gun classification dataset. To our knowledge this is the largest gun detection and classification dataset available for research and development purpose. We evaluate the gun detection performance on an edge device, and the gun/non-gun classification on the edge device or on the cloud server. 

A work in progress is developing the gun action recognition for the cloud server. This gun action recognition is expected to reduce the FPR further. 

\bibliographystyle{IEEEbib}
\bibliography{gun-dataset}

\end{document}